\pdfoutput=1
\documentclass[11pt]{article}

\usepackage{acl}
\usepackage{nth}
\usepackage{times}
\usepackage{tikz,pgfplots}
\usepackage{amssymb}
\usepackage{latexsym}
\usepackage{graphicx}
\usepackage{multirow}
\usepackage{array}
\usepackage{booktabs}

\usepackage[normalem]{ulem}
\useunder{\uline}{\ul}{}
\usepackage[T1]{fontenc}
\usepackage[utf8]{inputenc}

\usepackage{microtype}

\usepackage{inconsolata}

\title{\textsc{KnowShiftQA}: How Robust are RAG Systems when Textbook Knowledge Shifts in K-12 Education?}

\author{Tianshi Zheng\thanks{~~Equal Contribution}$^{\spadesuit}$, Weihan Li\footnotemark[1]$^{\clubsuit}$, Jiaxin Bai$^{\spadesuit}$, Weiqi Wang$^{\spadesuit}$, Yangqiu Song$^{\spadesuit}$ \\
  $^{\spadesuit}$Department of Computer Science and Engineering, HKUST, Hong Kong SAR, China\\
  $^{\clubsuit}$Department of Mechanical Engineering, The University of Tokyo, Tokyo, Japan\\
  \texttt{tzhengad@connect.ust.hk}\\
}
\begin{document}

\maketitle

\begin{abstract}
Retrieval-Augmented Generation (RAG) systems show remarkable potential as question answering tools in the K-12 Education domain, where knowledge is typically queried within the restricted scope of authoritative textbooks.
However, discrepancies between these textbooks and the parametric knowledge inherent in Large Language Models (LLMs) can undermine the effectiveness of RAG systems.
To systematically investigate RAG system robustness against such knowledge discrepancies, we introduce \textbf{\textsc{KnowShiftQA}}\footnote{\href{https://github.com/HKUST-KnowComp/KnowShiftQA}{https://github.com/HKUST-KnowComp/KnowShiftQA}}.
This novel question answering dataset simulates these discrepancies by applying deliberate hypothetical knowledge updates to both answers and source documents, reflecting how textbook knowledge can shift.
\textsc{KnowShiftQA} comprises 3,005 questions across five subjects, designed with a comprehensive question typology focusing on context utilization and knowledge integration.
Our extensive experiments on retrieval and question answering performance reveal that most RAG systems suffer a substantial performance drop when faced with these knowledge discrepancies.
Furthermore, questions requiring the integration of contextual (textbook) knowledge with parametric (LLM) knowledge pose a significant challenge to current LLMs. 
\end{abstract}
\begin{figure}[t]
    \centering
    \includegraphics[width=0.5\textwidth]{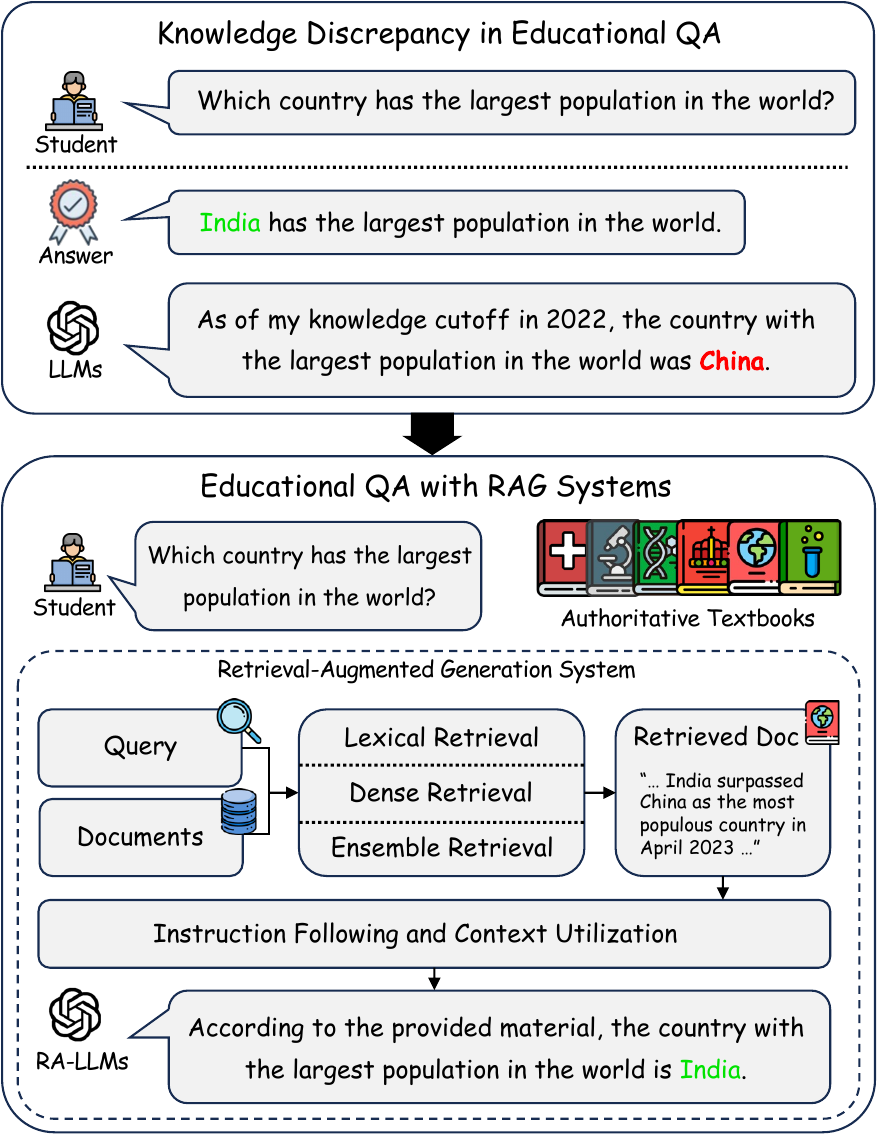}
    \caption{An illustration of knowledge discrepancy in educational QA and the application of RAG systems.}
    \label{fig:frame}
\end{figure}
% Please add the following required packages to your document preamble:
% \usepackage{booktabs}
% \usepackage{multirow}
\begin{table*}[]
\centering
\scriptsize
\begin{tabular}{ccc@{}}
\toprule
\textbf{Question Type} & \textbf{Reasoning Pattern and Example Question} & {\textbf{Hypothetical Knowledge Update}} \\ \midrule
\multirow{2}{*}{\raisebox{-2\height}{\textit{Simple Direct}}} & {\vspace{0.2cm}\raisebox{-0.3\height}{\includegraphics[height=0.4cm]{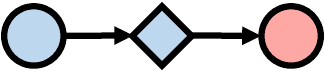}}}& {\begin{tabular}[c]{r}NV goggles - detect - \sout{Infrared} light\qquad\qquad\qquad\qquad\\ \textit{\textbf{\textcolor{red}{Ultraviolet}}} \qquad\qquad\qquad~~~~~~\end{tabular}\vspace{-0.2cm}}\\ \cmidrule(l){2-3} 
 & \parbox[]{6cm}{\vspace{0.0cm}What type of light is \textit{detected} by \textbf{night vision goggles}?} & {\begin{tabular}[c]{@{}l@{}}Original: Infrared Light\\ Updated: Ultraviolet Light~~~~~~~~~~~~~~~\end{tabular}} \\ \midrule
 
\multirow{2}{*}{\raisebox{-2\height}{\textit{Multi-hop Direct}}} & {\vspace{0.2cm}\raisebox{-0.3\height}{\includegraphics[height=0.4cm]{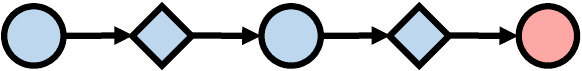}}} & {\begin{tabular}[c]{r}de Broglie eq. - developed by - \sout{Louis} de Broglie\\ \textit{\textbf{\textcolor{red}{Maurice}}} \quad
\quad\quad~
\end{tabular}\vspace{-0.2cm}} \\ \cmidrule(l){2-3} 
 & \parbox[]{6cm}{\vspace{0.0cm}Which scientist \textit{developed} an \textbf{equation} that can \textit{calculate} the \textbf{wavelength of a particle}?}  & {\begin{tabular}[c]{@{}l@{}}Original: Louis de Broglie\\ Updated: Maurice de Broglie~~~~~~~~~~~\end{tabular}} \\ \midrule
 
\multirow{2}{*}{\raisebox{-1.5\height}{\textit{Multi-hop Distant}}} & {\raisebox{-0.2\height}{\includegraphics[height=0.68cm]{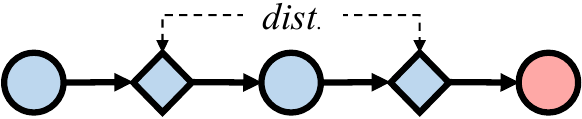}}} & {\begin{tabular}[c]{r}
\raisebox{0.1cm}{\raisebox{-0.0\height}{\sout{Na\textsuperscript{+}/K\textsuperscript{+}} Pump - creates - EC Gradient}}\qquad\qquad\vspace{-0.1cm}\\ \raisebox{0.0cm}{\textit{\textbf{\textcolor{red}{Ca\textsuperscript{2+}}}} \qquad
\qquad\qquad\qquad\qquad\qquad\qquad~~~~~~~~~~} \end{tabular}} \\ \cmidrule(l){2-3} 
 & \parbox[]{6cm}{\vspace{0.0cm} Which pump \textit{creates} an \textbf{electrochemical gradient} that \textit{enables} \textbf{secondary active transport} to occur?} & {\begin{tabular}[c]{@{}l@{}}Original: Sodium-potassium Pump\\ Updated: Calcium Pump\end{tabular}} \\ \midrule
 
\multirow{2}{*}{\raisebox{-2\height}{\textit{Multi-hop Implicit}}} & {\vspace{0.2cm}\raisebox{-0.3\height}{\includegraphics[height=0.4cm]{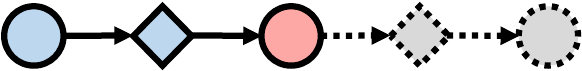}}} & {\begin{tabular}[c]{r}\sout{Polonium} - found in - Uranium ores\qquad\qquad\qquad\\ \textit{\textbf{\textcolor{red}{Thorium}}} \qquad\qquad\qquad
\qquad\qquad\qquad\qquad~~~\end{tabular}\vspace{-0.2cm}} \\ \cmidrule(l){2-3} 
 & \parbox[]{6cm}{\vspace{0.0cm} Who \textit{discovered} the \textbf{radioactive element} that is commonly \textit{found} in \textbf{uranium ores}?} & {\begin{tabular}[c]{@{}l@{}}Original: Marie Curie\\ Updated: Jöns Jacob Berzelius~~~~~~ \end{tabular}} \\ \midrule

\multirow{2}{*}{\raisebox{-1.5\height}{\textit{Distant Implicit}}} & {\raisebox{-0.2\height}{\includegraphics[height=0.68cm]{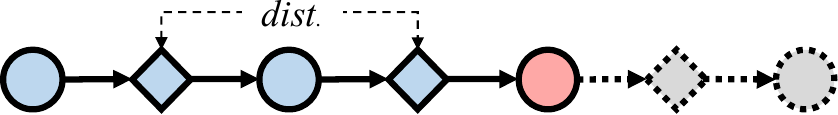}}\vspace{0.1cm}} & 
{\begin{tabular}[c]{r}
\raisebox{0.1cm}{\raisebox{-0.2\height}{\sout{Mitochondrion} - conducts - Cellular respiration}}\vspace{-0.1cm}\\ \textit{\textbf{\textcolor{red}{Golgi apparatus}}} \qquad\qquad\qquad\qquad\qquad~~~~~~~~~ 
\end{tabular}}  \\ \cmidrule(l){2-3}
 & \parbox[]{6cm}{\vspace{0.0cm} Who \textit{discovered} the \textbf{organelle} that is \textit{responsible} for the \textbf{biological process} that \textit{produces} \textbf{ATP}?} & {\begin{tabular}[c]{@{}l@{}}Original: Albert von Kölliker~~~~~~ \\ Updated: Camillo Golgi\end{tabular}} \\ \bottomrule
\multicolumn{3}{c}{{\raisebox{0.0\height}{\includegraphics[height=0.40cm]{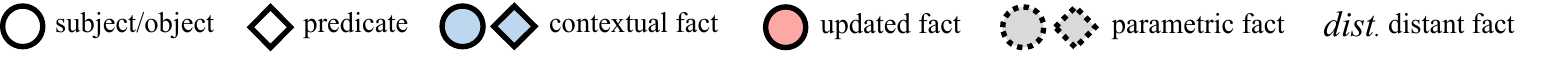}}\vspace{-0.2cm}}} 
\end{tabular}
\caption{Five question types in \textsc{KnowShiftQA} with their reasoning patterns, example questions, and hypothetical knowledge update illustrated in factual triplets and answers. In example questions, subjects and objects are marked in \textbf{bold}, while predicates are marked in \textit{italic}.}
\label{tab:typology}
\end{table*}
\section{Introduction}
In K-12 education, Question Answering (QA) systems serve as an important resource for learning assistance, where answers are precisely provided within a restricted knowledge scope from authoritative sources (i.e., textbooks) \cite{raamadhurai-etal-2019-curio,soares2021education}. Meanwhile, benefiting from the emergent abilities \cite{wei2022emergent} of Large Language Models (LLMs) and advanced information retrieval (IR) methods, Retrieval-Augmented Generation (RAG) systems have achieved remarkable performance in various knowledge-intensive tasks in natural language processing \cite{lewis2020retrieval,jiang2023active,gao2024retrievalaugmented}, demonstrating their great potential as QA systems in K-12 education \cite{gan2023large,KASNECI2023102274,yan2024practical}.

In K-12 educational QA, one of the primary concerns is ensuring that the knowledge conveyed in the answer is consistent with the officially designated textbooks \cite{Extance2023ChatGPTClassroom}. However, notable knowledge shifts (i.e. factuality discrepancies) may exist between the knowledge in textbooks and the internal knowledge of LLMs, due to the evolving nature of facts \cite{arbesman2012half}, updates in pedagogical approaches \cite{provenzo2011textbook}, as well as regional and cultural variations in content \cite{patel2015decolonizing}. It remains unclear whether RAG systems can robustly incorporate knowledge from authoritative sources and generate consistent answers under such knowledge discrepancies (scenario illustrated in Figure \ref{fig:frame}).

To fill this gap, we aim to systematically assess the robustness of RAG systems in performing question answering in K-12 education when encountering knowledge shifts. We present \textbf{\textsc{KnowShiftQA}}, a new dataset containing 3,005 multiple-choice questions covering the subjects of Physics, Chemistry, Biology, Geography, and History from the middle-school curriculum. To simulate the factuality discrepancy between LLMs and textbooks, we conduct a \textit{hypothetical knowledge update}, in which we modify the original factual knowledge from the textbooks into plausible alternatives while maintaining coherent and consistent context. Moreover, we tailored a comprehensive question typology to stress-test the \textit{context utilization} and \textit{knowledge integration} abilities of LLMs under such scenarios.

We conducted extensive experiments with various retrieval methods and LLMs.
Our findings indicate that most RAG systems exhibit a considerable performance degradation under knowledge shifts.
Notably, while many LLMs can effectively incorporate distant contextual facts, they struggle to seamlessly integrate their parametric knowledge with this contextual information.
In terms of retrieval, traditional lexical-based methods demonstrate advantages due to the specificity of academic terms; their performance can be further enhanced through an ensemble re-ranking mechanism.
Supervised retrieval methods (e.g., Contriever), when fine-tuned directly on our dataset, yield superior performance.
Overall, our results highlight the fragility of current RAG systems under knowledge shifts and the challenges LLMs face in effectively combining contextual and parametric knowledge for complex question answering.

\section{The \textsc{KnowShiftQA} Dataset}
 In this section, we introduce the methodology of hypothetical knowledge update and the design of our question typology. The curation pipeline and detailed statistics of the \textsc{KnowShiftQA} dataset are provided in Appendix \ref{app:pipeline} and \ref{app:data_stats}, respectively.

\subsection{Hypothetical Knowledge Update}
One of the core objectives of our dataset is to simulate the knowledge discrepancy between LLMs and authoritative textbooks when performing educational QA. However, such discrepancies are often fuzzy and highly sparse in real-world data, making it infeasible to collect and organize. Consequently, we designed the methodology of hypothetical knowledge update, performing it on high-quality open-source textbooks. The general procedures are as follows: 1) Curate factual questions  from textbook paragraphs following our designed question typology. 2) Select a plausible but factually incorrect answer as the updated ground-truth answer. 3) Replace all occurrences of the original answer in the paragraph with the updated answer and adjusted other relevant statements in the context to ensure that the updated paragraph is coherent and consistent. This process is further guaranteed through extensive human curation and verification. Examples of hypothetical knowledge updates in our dataset are provided in Appendix \ref{app:bench_examples}.

\subsection{Question Typology}
We identify two potential challenges for LLMs when performing QA under knowledge shifts: 1) \textit{Context Utilization}: Whether LLMs can identify and utilize the corresponding facts from the context; and 2) \textit{Knowledge Integration}: Whether LLMs can incorporate their own parametric knowledge with contextual facts in question answering. To this end, we designed our question typology, as illustrated in Table~\ref{tab:typology}, to investigate such abilities. The two basic question types, \textit{Simple Direct} and \textit{Multi-hop Direct}, aim to evaluate LLMs on simple factual recall and multi-hop reasoning. Based on these two types, we developed the \textit{Multi-hop Distant} type to evaluate the context utilization ability for distant facts from the passage, and the \textit{Multi-hop Implicit}\footnote{The word 'implicit' indicates that the questions indirectly query the updated facts by embedding them within the middle of the multi-hop reasoning chain.} type to evaluate the knowledge integration ability that combines their own factual knowledge with retrieved facts. Moreover, the \textit{Distant Implicit} type poses a greater challenge by combining both features. To ensure our evaluation of knowledge integration ability is independent of knowledge coverage, we restrict the facts requiring LLMs' own knowledge to be head-knowledge only \cite{sun-etal-2024-head}.
%We aim to create questions with different reasoning patterns to enrich the complexity coverage in our dataset. Inspired by studies in complex logical reasoning over knowledge graphs \cite{hamilton2019embedding, ren2020query2box, DBLP:journals/corr/abs-2302-13114}, we designed an ontology-aware typology that includes six reasoning patterns in questions (\textit{1a}, \textit{1p}, \textit{pc}, \textit{pa}, \textit{2p}, and \textit{2i}), detailed in Appendix \ref{app:typology_details}. For instance, question type \textit{pc} queries the \textit{objective} from an \textit{s-p-o} triplet with a corresponding \textit{condition} as a constraint of the predicate. In the question \textit{'Which scientific principle did Gregor Mendel establish in 1865?'}, we identify \textit{'Gregor Mendel'} as the subject, \textit{'establish'} as the predicate, \textit{'in 1865'} as a condition constraining the predicate \textit{'establish'}, and \textit{'Which scientific principle'} as the objective the question seeks to identify. As statements with complex structure (e.g. multi-hop) are less frequent in natural language corpus \cite{rajpurkar-etal-2016-squad}, more sophisticated reasoning patterns, such as 3-hop projection, are not included in our typology. 

\section{Experiments and Analyses}
Typically, RAG systems first conduct document retrieval based on given queries, then perform question answering with LLMs based on the retrieved information loaded in the context. In this section, we comprehensively evaluate and analyze the performance of retrieval methods and LLMs on the \textsc{KnowShiftQA} dataset. For details of all tested methods and models, please refer to Appendix \ref{app:models}.
\begin{table}[t]
\hfill
\centering
\scriptsize
\begin{tabular}{lccc}
\toprule
\multicolumn{1}{c}{\textbf{Retrieval Methods}} & \textbf{Category} & \textbf{R@1} & \textbf{R@5} \\ \midrule
TF-IDF \cite{sparck1972statistical} & Lexical & 65.82 & 88.72 \\
BM25 \cite{robertson1994okapi} & Lexical & {82.73} & 95.27 \\
SPLADE \cite{formal2021spladev2sparselexical}& Lexical/Dense & 78.04 & 90.12 \\
Contriever \cite{izacard2022unsupervised}& Dense & 53.18 & 81.80 \\
Con.-msmarco \cite{izacard2022unsupervised} & Dense & 76.17 & 93.54 \\
Mistral-embed \cite{MistralAIEmbeddings} & Dense & 78.74 & {95.31} \\
Ada-002 \cite{OpenAI2022Embedding} & Dense & {79.23} & {95.44} \\
Query Rewrite \cite{ma2023queryrewritingretrievalaugmentedlarge} & Pre-Retrieval & 78.87 & 94.21 \\
Hybrid Rerank (BM25 + Ada-002) & Ensemble & {\ul 84.43} & {\ul 96.04} \\ 
Contriever (fine-tuned) & Dense+FT & {\ul 84.19} & {\ul 98.96} \\
Con.-msmarco (fine-tuned) & Dense+FT & \textbf{87.95} & \textbf{99.50} \\

\bottomrule
\end{tabular}
\caption{Document retrieval performances (in recall @1/5 \%) of retrieval methods from different categories in the \textsc{KnowShiftQA} dataset. The highest scores are marked as \textbf{bold}, while the 2\textsuperscript{nd} and 3\textsuperscript{rd}-best scores are \underline{underlined}. The retrieval granularity is set to paragraph.}
\label{tab:retrieval}
\end{table}
\subsection{Retrieval Performance}
Experimental results of the retrieval methods are presented in Table \ref{tab:retrieval}. Traditional lexical retrieval methods, such as BM25, demonstrated strong performance on our dataset, while dense retrieval methods, such as Mistral-embed and Ada-002, achieved comparable performance. Since our dataset focuses on the K-12 education domain, lexical retrieval effectively captures domain-specific keywords in the queries and identifies the corresponding  documents. This characteristic underscores the need for fine-tuning dense retrieval models on our dataset. To this end, we fine-tuned both Contriever and Contriever-msmarco on our documents, resulting in significant improvements and highlighting the importance of corpus-specific fine-tuning in educational document retrieval.

Moreover, ensemble methods have demonstrated their effectiveness in various information retrieval tasks \cite{thakur2021beir}. We implemented a hybrid approach that retrieves the top  \textit{k}\footnote{We use \textit{k} = 6 as the optimal hyperparameter setting.} documents with Ada-002, followed by re-ranking with BM25. This ensemble method marginally improved the retrieval performance in both metrics. Conversely, query rewriting\footnote{We use \texttt{Mistral-small-2409} as the LLM query rewriter.} \cite{ma2023queryrewritingretrievalaugmentedlarge} implemented upon BM25 did not yield performance gains. Detailed results across subjects and question types are provided in Appendix \ref{app:result_details}.
\begin{table*}[h]
\scriptsize
\centering
\begin{tabular}{lcccccc}
\toprule
\multicolumn{1}{c}{\multirow{2}{*}{\textbf{Large Language Models}}} & \multicolumn{5}{c}{\textbf{Question Typology}} & \multirow{2}{*}{\textbf{Average}} \\ \cmidrule(){2-6}
\multicolumn{1}{c}{} & \textit{Simple Direct} & \textit{Multi-hop Direct} & \textit{Multi-hop Distant} & \textit{Multi-hop Implicit} & \textit{Distant Implicit} &  \\ \midrule
Mistral-7b \cite{mistral2023mistral7b} & 77.70 & 69.31 & 72.74 & 45.32 & 33.98 & 61.26 \\
Mixtral-8x22b \cite{mistral2023mixtralofexperts} & 84.10 & 84.58 & 87.15 & 73.86 & 65.37 & 79.67 \\
Mistral-small-2409 \cite{mistralai2024september} & 87.87 & 88.70 & 89.69 & 72.18 & 60.81 & 80.77 \\
Mistral-large-2407 \cite{mistralai2024large2} & 83.93 & 83.51 & 87.29 & {82.25} & 70.57 & 81.66 \\ \midrule
Gemini-1.5-flash \cite{googledeepwind2024geminiflash}& 80.98 & 82.44 & 87.57 & 76.02 & 63.25 & 78.54 \\
Gemini-1.5-pro \cite{gemini_pro_2024} & 86.56 & 87.63 & 88.42 & 76.26 & 63.58 & 81.10 \\ \midrule
Llama3-8b \cite{meta_llama_3_2024} & 90.33 & 88.85 & 89.69 & 63.55 & 49.43 & 77.77 \\
Llama3-70b \cite{meta_llama_3_2024} & {\ul 96.72} & {\ul96.49} & {\ul96.89} & 79.14 & 63.25 & {87.42} \\ \midrule
GPT-3.5-turbo \cite{OpenAI2023ChatGPT} & 92.62 & 90.53 & 91.24 & 71.22 & 57.72 & 81.73 \\
GPT-4 \cite{openai2023gpt4} & 89.51 & 89.47 & 90.68 & 78.66 & {70.89} & 84.46 \\
GPT-4-turbo \cite{openai2024devday} & {95.74} & {96.18} & {96.19} & 81.06 & {71.71} & {88.99} \\
GPT-4o \cite{openai2024gpt4o} & 91.97 & 94.81 & 93.64 & {81.29} & 70.41 & 87.09 \\ \midrule
Claude-3-sonnet \cite{anthropic2024a} & 94.59 & 92.82 & 93.64 & 77.70 & 60.65 & 84.69 \\
Claude-3.5-sonnet \cite{anthropic2024b} & \textbf{97.54} & {\ul96.49} & {95.62} & {\ul83.69} & {\ul73.82} & {\ul90.08}\\\midrule
o1-mini \cite{openai2024o1} & {\ul 95.90} & 95.73 & {\ul 97.03} & {\ul 85.85} & {\ul 75.45} & {\ul 90.55} \\
o1-preview \cite{openai2024o1} & 95.08 & \textbf{97.71} & \textbf{97.46} & \textbf{86.33} & \textbf{78.86} & \textbf{91.68} \\ 
\bottomrule
\end{tabular}
\caption{Question answering performances (in accuracy \%) of LLMs in the \textsc{KnowShiftQA} benchmark with corresponding documents provided. The highest scores are marked as \textbf{bold}, while the 2\textsuperscript{nd} and 3\textsuperscript{rd}-best scores are \underline{underlined}. All LLMs are tested under zero-shot settings, with a \textit{Locate-and-Answer} prompting approach that facilitates active information acquisition from contextual documents, with details in Appendix \ref{app:prompting_approach}.}
\label{tab:llm_performance}
\end{table*}

\subsection{Question Answering Performance}

Experimental results of LLMs in question answering are presented in Table \ref{tab:llm_performance}. Most LLMs exhibited comparable performance for question types \textit{Simple Direct}, \textit{Multi-hop Direct}, and \textit{Multi-hop Distant}. This suggests that \textbf{both multi-hop reasoning and distant context utilization do not pose significant challenges for modern LLMs}. However, for \textit{Multi-hop Implicit} questions, which requires the integration of contextual and internal knowledge, a substantial performance disparity emerges between smaller open-source LLMs and advanced models. This disparity is further amplified for \textit{Distant Implicit} questions, where the reasoning chains are getting more complexed. The most capable model we tested, o1-preview, attains over 80\% accuracy on \textit{Implicit} questions, whereas Mistral-7b's performance falls below 40\%. These findings indicate that \textbf{context-memory knowledge integration is an emergent capability presenting greater difficulties for LLMs under knowledge discrepancies}, particularly when coupled with the need for distant context utilization.
\begin{table}[]
\centering
\scriptsize
\begin{tabular}{lccc}
\toprule
\multicolumn{1}{c}{\multirow{2}{*}{\textbf{RAG System}}} & \multicolumn{2}{c}{\textbf{Hypothetical Knowledge Update}} & \multirow{2}{*}{\textbf{Drop}} \\ \cmidrule{2-3}
\multicolumn{1}{c}{} & Before & After &  \\ \midrule
Llama3-8b + Ada-002 & 87.49 & 62.60 & 24.89 \\
Llama3-8b + Rerank & 88.49 & 66.02 & 22.47 \\
GPT-4o + Ada-002 & 96.57 & 69.65 & 26.92 \\
GPT-4o + Rerank & 97.10 & 73.71 & 23.39 \\ \bottomrule
\end{tabular}
\caption{Performance drop of RAG systems with hypothetical knowledge updates in our benchmark.}
\label{tab:drop}
\end{table}
\subsection{Overall Performance}
How do knowledge shifts affect the performance of RAG systems in educational question-answering applications?
To answer this question, we selected two representative LLMs: Llama3-8b from open-source models and GPT-4o from proprietary models.
These were combined with two high-performing retrieval methods: Ada-002 and hybrid rerank.
The resulting RAG systems were tested on questions \textit{before} and \textit{after} hypothetical knowledge updates, with the results presented in Table~\ref{tab:drop}.
We observed a significant accuracy drop of 22--27\% after these knowledge shifts, indicating substantial performance degradation in modern RAG systems when faced with such scenarios.

\section{Related Work}
\paragraph{Retrieval-Augmented Generation}
Following the categorization by \citet{gao2024retrievalaugmented}, the RAG methods employed in our experiment fall into the categories of \textit{Naive RAG} (lexical, dense) and \textit{Advanced RAG} (rerank, rewrite). Recently, \textit{Modular RAG} has emerged to enhance the adaptability and versatility of RAG systems \cite{shao2023enhancing,asai2023selfrag}. Furthermore, recent advances in vector databases \cite{han2023comprehensivesurveyvectordatabase} and neural graph databases \cite{bai2025challengesagenticneuralgraph} highlight the potential for adaptive and generalizable RAG search with structured data \cite{deng2024texttupletableinformationintegrationtexttotable, bai2025autoschemakgautonomousknowledgegraph}.
\newpage
\paragraph{Educational Question Answering}
Prior to the emergence of RAG systems utilizing LLMs, various educational QA systems were developed to provide pedagogically appropriate responses to student inquiries \cite{abdi2018qapd,8683538}. While recent literature explores LLM applications in QA and learning assistance roles \cite{nye2023generative,kuo2023leveraging,wang2024large}, it is also crucial to leverage logical inference abilities, such as abductive reasoning \cite{bai2024advancingabductivereasoningknowledge,yim2024evaluatingenhancingllmsagent,zheng2025logidynamicsunravelingdynamicslogical} and inductive reasoning \cite{qiu2024phenomenal,li2025patternsprinciplesfragilityinductive,zheng2025cursecotlimitationschainofthought}, for understanding students' learning outcomes in response to their queries.
\paragraph{Knowledge Discrepancy in LLMs}
Mitigating knowledge discrepancies or conflicts within LLM applications is a fundamental challenge in LLM research \cite{xu2024knowledgeconflictsllmssurvey}. Researchers have proposed tuning-based \cite{li2022large} and prompting-based \cite{zhou2023contextfaithfulpromptinglargelanguage} methods to enhance LLMs' robustness to such conflicts. However, the implications of these conflicts in educational applications and RAG systems remain relatively underexplored. Moreover, it is important to consider the trade-off between factuality robustness \cite{yu2024kolacarefullybenchmarkingworld,zong2024comparisonqaevaluatingfactualityrobustness,zheng2025automationautonomysurveylarge} and flexibility in instruction following when such discrepancies are present.

\section{Conclusion}
This paper systematically evaluates the robustness of RAG systems in K-12 educational question answering under knowledge discrepancies using a comprehensive dataset \textsc{KnowShiftQA}. Experimental findings reveal substantial performance degradation in RAG systems when faced with knowledge discrepancies, which is primarily attributed to deficiencies in incorporating contextual and parametric knowledge in question answering—an emergent and challenging ability for modern large language models.

\section*{Limitations}
We discuss three main limitations of our work.

First, \textsc{KnowShiftQA} employs the approach of hypothetical knowledge updates to effectively simulate real-world knowledge discrepancies for two primary reasons: (1) Real-world knowledge conflicts are often sparse, noisy, and difficult to systematically collect or organize into a cohesive dataset suitable for large-scale evaluation. Hypothetical updates provide a scalable and controlled alternative, allowing us to bypass these limitations. (2) By leveraging a systematic annotation and curation pipeline, we can generate questions with diverse and well-defined reasoning patterns that align with our typology, enabling more robust evaluation of complex question-answering tasks. For future research centered on real-world knowledge discrepancies, we recommend an alternative methodology that incorporates temporal attributes \cite{chen2021datasetansweringtimesensitivequestions, zhang2024mitigatingtemporalmisalignmentdiscarding}. This approach focuses on identifying outdated facts and capturing time-sensitive data (e.g., economic trends, annual events, or societal changes) to construct datasets that reflect real-world knowledge updates. While promising, this method is constrained by the limited overlap between time-sensitive data—often numerical or attribute-specific—and the broader contextual needs of educational question-answering tasks, which may reduce the comprehensiveness of the resulting datasets.

Next, regarding document retrieval, some recent hierarchical retrieval paradigms, such as GraphRAG \cite{edge2024localglobalgraphrag} and HippoRAG \cite{gutiérrez2024hipporagneurobiologicallyinspiredlongterm}, are not included in our experiments due to their implementation complexity. However, we believe that such structured paradigms could effectively enhance retrieval performance in our scenario, as educational documents are well-structured and contain high-quality factual knowledge.

Finally, this paper primarily evaluates the robustness of RAG systems in the proposed scenario, with experiments conducted using various retrieval methods and large language models. Potential improvements could be achieved through the design of tailored reasoning frameworks via prompting, in-context learning or alignment in LLMs, which we leave for future research.

\section*{Ethics Statement}
In constructing the \textsc{KnowShiftQA} dataset, we collected text from open-access textbooks. Detailed sources and licenses are provided in Appendix \ref{app:source_textbooks}. The human curation and verification in our annotation pipeline were carried out by a group of postgraduate students with extensive experience in NLP research. We ensured that the updated knowledge is free from harmful or toxic content. It is important to note that our dataset is designed solely to evaluate the robustness of Retrieval-Augmented Generation systems under scenarios with knowledge discrepancies and is not suitable for assessing the factual accuracy of QA systems.
% Entries for the entire Anthology, followed by custom entries

\section*{Acknowledgments}
The authors of this paper were supported by the ITSP Platform Research Project (ITS/189/23FP) from ITC of Hong Kong, SAR, China, and the AoE (AoE/E-601/24-N), the RIF (R6021-20) and the GRF (16205322) from RGC of Hong Kong SAR, China. 
\bibliography{main}

\newpage
\onecolumn
\appendix
\section{Source Textbooks}
\label{app:source_textbooks}
\begin{figure}[h]
    \centering
    \includegraphics[width=0.5\textwidth]{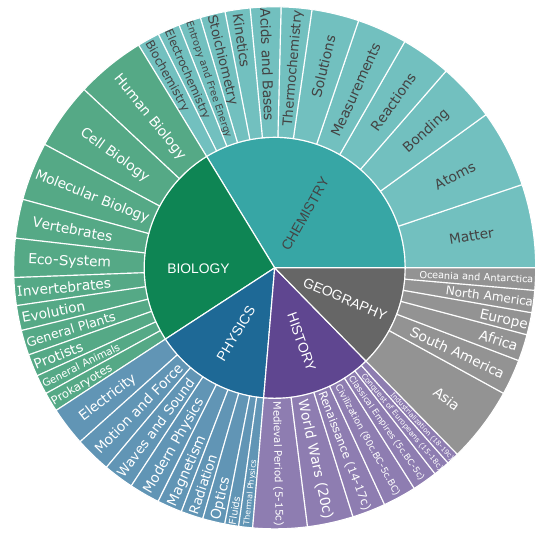}
    \caption{Distribution of five subjects and their corresponding topics included in the \textsc{KnowShiftQA} dataset.}
    \label{fig:topics}
\end{figure}

The documents in the \textsc{KnowShiftQA} dataset is organized based on the following public textbooks (the detailed topics included in each subjects are illustrated in Figure \ref{fig:topics}.):
\paragraph{Physics}\textit{Physics} by Openstax (CC-BY-4.0\footnote{https://creativecommons.org/licenses/by/4.0/deed.en}) \url{https://openstax.org/details/books/physics}
\paragraph{Chemistry}\textit{Chemistry} by Openstax (CC-BY-4.0\footnotemark[1])
\url{https://openstax.org/details/books/chemistry-2e}
\paragraph{Biology}\textit{Biology} by Openstax (CC-BY-4.0\footnotemark[1])
\url{https://openstax.org/details/books/biology-2e}
\paragraph{History}\textit{World History} by OER Commons (CC-BY-NC-4.0\footnote{https://creativecommons.org/licenses/by-nc/4.0/deed.en}) \url{https://oercommons.org/courses/world-history-2} 
\paragraph{Geography}\textit{World Regional Geography} by Sailor Academy (CC-BY-3.0\footnote{https://creativecommons.org/licenses/by/3.0/})
\url{https://learn.saylor.org/course/view.php?id=722}\\

\newpage

\section{Curation Pipeline}
\label{app:pipeline}
\begin{figure*}[h]
    \centering
    \includegraphics[width=1\textwidth]{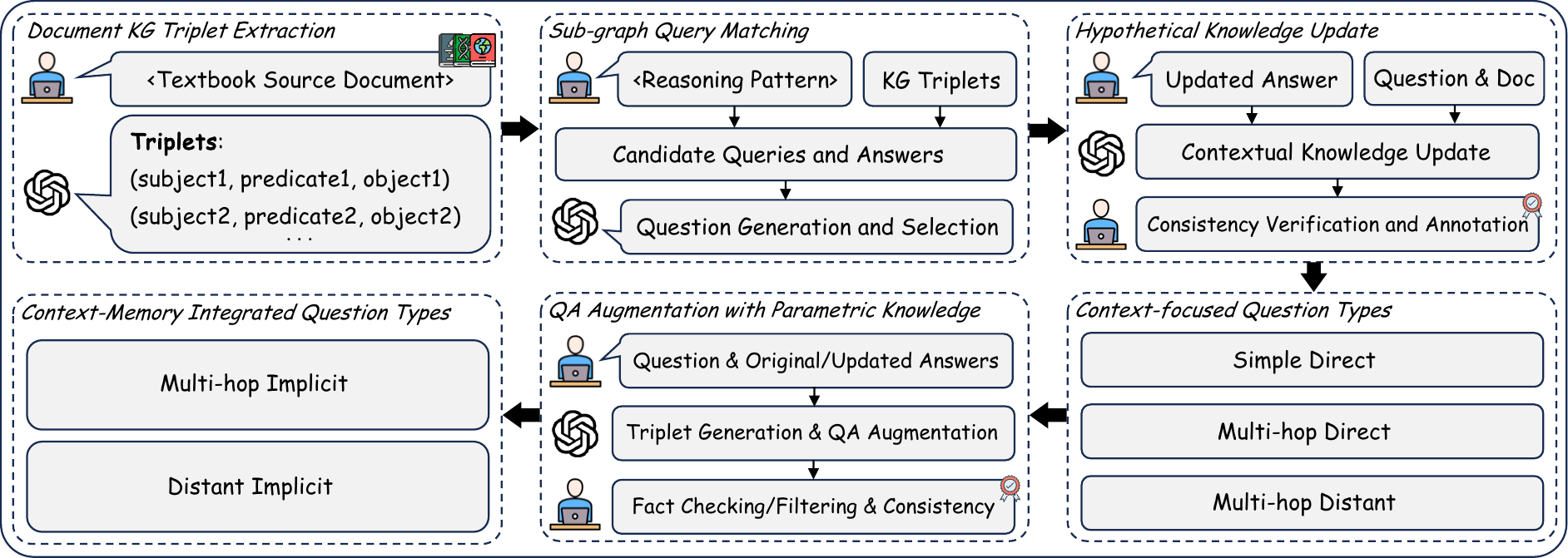}
    \caption{An overview of the data curation pipeline of the \textsc{KnowShiftQA} dataset.}
    \vspace{-10pt}
    \label{fig:pipeline}
\end{figure*}

The curation pipeline of our dataset is illustrated in Figure \ref{fig:pipeline}. We first perform triplet extraction on the textbook documents and generate a document-level knowledge graph (KG). Next, we perform sub-graph matching based on fixed reasoning patterns to sample candidate queries, and selectively transform them into natural language questions. Then, hypothetical knowledge update is executed and verified to guarantee consistency between the updated answer and the document. 

Context-focused question types, including \textit{Simple Direct}, \textit{Multi-hop Direct}, and \textit{Multi-hop Distant}, are acquired through this process. For \textit{Multi-hop Distant} questions, we leverage distant facts, defined as connected triplet pairs that are separated in the document's sequential ordering. These questions are only assigned to documents containing no fewer than 200 words. To generate the other two \textit{Implicit} question types that require knowledge integration, we perform extra QA augmentation followed by an expert verification process. The design of our question typology is motivated by studies in complex logical reasoning \cite{bai2023sequentialqueryencodingcomplex,zheng2024clrfactevaluatingcomplexlogical,zheng2025enhancingtransformersgeneralizablefirstorder}.

Our data curation process is performed through an integrated framework involving both human annotators and LLMs. For LLM annotation, we adopted \texttt{Claude-3.5-Sonnet} for its outstanding instruction-following ability. Following manual verification, 90.5\% of these queries were retained or underwent minor refinements to become high-quality questions, yielding an overall success rate of 86.4\%. The total API cost for data annotation is approximately 300 USD.
\newpage

\section{Model Details}
\label{app:models}
In this section, we briefly introduce all tested retrieval methods and large language models in our benchmark experiment.

In the retrieval stage of our experiments, we employ a diverse range of retrieval methods. For traditional lexical retrieval, we include \textbf{TF-IDF} \cite{sparck1972statistical}, which vectorizes documents and queries based on term frequency and inverse document frequency, and \textbf{BM25} \cite{robertson1994okapi}, which enhances TF-IDF with document length normalization and probabilistic term weighting. We also incorporate \textbf{SPLADE} \cite{formal2021spladev2sparselexical}, a method that bridges lexical and dense retrieval paradigms. For dense retrieval, we evaluate several methods that encode questions and documents into the same vector space: \textbf{Contriever} \cite{izacard2022unsupervised}, an unsupervised text encoder; its fine-tuned variant \textbf{Contriever-msmarco}, which we further enhanced by applying contrastive learning \cite{izacard2022unsupervised} to fine-tune both models on the \textsc{KnowShiftQA} dataset for improved retrieval capability; and two closed-source embedding models, \textbf{Mistral-embed} \cite{MistralAIEmbeddings} and \textbf{Ada-002} \cite{OpenAI2022Embedding}.  Additionally, we explore a pre-retrieval method, \textbf{Query Rewrite} \cite{ma2023queryrewritingretrievalaugmentedlarge}, which reformulates queries to improve retrieval performance. Finally, we implement a \textbf{Hybrid Rerank} approach that combines BM25 and Ada-002, leveraging the strengths of both lexical and dense retrieval methods.

In our evaluation, we employ a diverse set of state-of-the-art large language models to assess their performance across various tasks. The models include: Mistral AI's open-source models, ranging from the compact \textbf{Mistral-7b} \cite{mistral2023mistral7b} to the more advanced MoE model \textbf{Mixtral-8x22b} \cite{mistral2023mixtralofexperts}, and their latest iterations \textbf{Mistral-small-2409} \cite{mistralai2024september} and \textbf{Mistral-large-2407} \cite{mistralai2024large2}. Google's Gemini models are represented by \textbf{Gemini-1.5-flash} \cite{googledeepwind2024geminiflash} and \textbf{Gemini-1.5-pro} \cite{gemini_pro_2024}. Meta's \textbf{Llama3} series is included with both 8b and 70b parameter versions \cite{meta_llama_3_2024}. We also evaluate OpenAI's models, including \textbf{GPT-3.5-turbo} \cite{OpenAI2023ChatGPT}, \textbf{GPT-4} \cite{openai2023gpt4}, \textbf{GPT-4-turbo} \cite{openai2024devday}, and \textbf{GPT-4o} \cite{openai2024gpt4o}. Anthropic's LLMs are represented by \textbf{Claude-3-sonnet} \cite{anthropic2024a} and \textbf{Claude-3.5-sonnet} \cite{anthropic2024b}. Lastly, we include OpenAI's latest o1 series, \textbf{o1-mini} and \textbf{o1-preview} \cite{openai2024o1}, which achieve remarkable performance across various metrics through inference-time scaling. This comprehensive selection allows us to compare a wide range of model architectures and sizes, providing valuable insights into the current state of LLM capabilities.

\newpage
\section{Dataset Example}
\label{app:bench_examples}
We provide an example question and its corresponding paragraph before and after hypothetical knowledge update in Table \ref{tab:example_question}.
\begin{table*}[h]
\centering
\small
\begin{tabular}{@{}c|clll@{}}
\toprule
\multirow{5}{*}{\textbf{Original Paragraph}} & \multicolumn{4}{c}{\multirow{5}{*}{\begin{tabular}[c]{@{}c@{}}... The halophiles, which means "salt-loving", \\ live in environments with high levels of salt. \\ They have been identified in the Great Salt Lake in Utah \\ and in the Dead Sea between Israel and Jordan, which have \\ salt concentrations several times that of the oceans. ...\end{tabular}}} \\
 & \multicolumn{4}{c}{} \\
 & \multicolumn{4}{c}{} \\
 & \multicolumn{4}{c}{} \\
 & \multicolumn{4}{c}{} \\ \midrule
\multirow{5}{*}{\textbf{Updated Paragraph}} & \multicolumn{4}{c}{\multirow{5}{*}{\begin{tabular}[c]{@{}c@{}}... The halophiles, which means "\textit{\textbf{\textcolor{red}{pressure}}}-loving", \\ live in environments with high levels of \textit{\textbf{\textcolor{red}{pressure}}}. \\ They have been identified in the \textit{\textbf{\textcolor{red}{Mariana Trench}}}\\  in \textit{\textbf{\textcolor{red}{western Pacific Ocean}}}, which have \\ higher \textit{\textbf{\textcolor{red}{pressure}}} than other environments. ...\end{tabular}}} \\
 & \multicolumn{4}{c}{} \\
 & \multicolumn{4}{c}{} \\
 & \multicolumn{4}{c}{} \\
 & \multicolumn{4}{c}{} \\ \midrule
\multirow{5}{*}{\textbf{Question}} & \multicolumn{4}{c}{\multirow{5}{*}{\begin{tabular}[c]{@{}l@{}}In which type of environments do halophiles typically live?\\ A. High Acidity Environments.\\ B. High Salt Environments.\\ C. High Pressure Environments.\\ D. High Sugar Environments.\end{tabular}}} \\
 & \multicolumn{4}{c}{} \\
 & \multicolumn{4}{c}{} \\
 & \multicolumn{4}{c}{} \\
 & \multicolumn{4}{c}{} \\ \midrule
\textbf{Original Answer} & \multicolumn{4}{c}{B. High Salt Environments.} \\ \midrule
\textbf{Updated Answer} & \multicolumn{4}{c}{\textit{\textbf{\textcolor{red}{C. High Pressure Environments.}}}} \\ \bottomrule
\end{tabular}
\caption{An example of hypothetical knowledge update for a question in Biology. The modifications of factual knowledge and contextual information in the paragraph are highlighted in \textit{\textbf{\textcolor{red}{red}}}.}
\label{tab:example_question}
\end{table*}

\newpage

\section{Dataset Statistics}
The statistics for the documents and questions in the \textsc{KnowShiftQA} dataset are provided in Table \ref{tab:doc_stats} and Table \ref{tab:ques_stats}, respectively.
\label{app:data_stats}
% Please add the following required packages to your document preamble:
% \usepackage{booktabs}
% \usepackage{multirow}
\begin{table*}[h]
\centering
\small
\begin{tabular}{lcccccc}
\toprule
\multicolumn{1}{c}{\multirow{2}{*}{\textbf{Document Statistics}}} & \multicolumn{5}{c}{\textbf{Subjects}} & \multirow{2}{*}{\textbf{Total}} \\ \cmidrule(lr){2-6}
\multicolumn{1}{c}{} & \textit{Chem.} & \textit{Bio.} & \textit{Phys.} & \textit{Geo.} & \textit{Hist.} &  \\ \midrule
Average Document Length & 315.5 & 588.1 & 647.3 & 429.8 & 277.1 & 437.3 \\
Num of Documents & 291 & 671 & 166 & 471 & 606 & 2205 \\ \bottomrule
\end{tabular}
\caption{Average length (in word counts) and quantities of documents in different subjects.}
\label{tab:doc_stats}
\end{table*}

\begin{table*}[h]
\centering
\scriptsize
\begin{tabular}{lcccccc}
\toprule
\multicolumn{1}{c}{\multirow{2}{*}{\textbf{\# Questions (avg. length)}}} & \multicolumn{5}{c}{\textbf{Question Types}} & \multirow{2}{*}{\textbf{Total}} \\ \cmidrule(){2-6}
\multicolumn{1}{c}{} & \textit{Simple Direct} & \textit{Multi-hop Direct} & \textit{Multi-hop Distant} & \textit{Multi-hop Implicit} & \textit{Distant Implicit} &  \\ \midrule
Chemistry & 73 (12.4) & 88 (17.8) & 78 (17.9) & 40 (17.4) & 68 (21.1) & 347 (17.3) \\
Biology & 148 (11.1) & 170 (16.7) & 248 (16.7) & 94 (16.7) & 213 (20.7) & 873 (16.7) \\
Physics & 46 (12.0) & 49 (16.7) & 49 (15.6) & 41 (16.9) & 39 (19.9) & 224 (16.1) \\
Geography & 141 (12.1) & 147 (16.7) & 162 (16.8) & 96 (17.7) & 144 (21.3) & 690 (16.9) \\
History & 202 (12.8) & 201 (16.1) & 171 (17.0) & 146 (18.1) & 151 (21.6) & 871 (16.8) \\ \midrule
\multicolumn{1}{c}{\textbf{Total}} & 610 (12.1) & 655 (16.7) & 708 (16.9) & 417 (17.5) & 615 (21.0) & 3005 (16.8) \\ \bottomrule
\end{tabular}
\caption{Average question length (in word counts) and quantities of questions in different subjects and types.}
\label{tab:ques_stats}
\end{table*}
\section{Result Details}
\label{app:result_details}

The comprehensive evaluation results for retrieval approaches and large language models across various subjects and question types are presented in Table \ref{tab:retrieval_result} and Table \ref{tab:llm_full_result}, respectively. We observed that the performance of lexical retrieval methods correlates positively with question length, while dense retrieval methods exhibit an inverse relationship. This finding suggests that there is potential for developing more sophisticated ensemble methodologies that could fully leverage the strengths of both approaches.

\begin{table*}[h]
\centering
\scriptsize
\begin{tabular}{lccccccccccc}
\toprule
\multicolumn{1}{c}{\multirow{2}{*}{\textbf{Retrieval Methods}}} & \multicolumn{5}{c}{\textbf{Subjects}} & \multicolumn{5}{c}{\textbf{Question Types}} & \multirow{2}{*}{\textbf{Average}} \\ \cmidrule(lr){2-11}
\multicolumn{1}{c}{} & \textit{chem.} & \textit{bio.} & \textit{phys.} & \textit{geo.} & \textit{hist.} & \multicolumn{1}{l}{\textit{Sim. Dir.}} & \multicolumn{1}{l}{\textit{Mul. Dir.}} & \multicolumn{1}{l}{\textit{Mul. Dis.}} & \multicolumn{1}{l}{\textit{Mul. Imp.}} & \multicolumn{1}{l}{\textit{Dis. Imp.}} &  \\ \midrule
BM25 & 87.90 & 84.65 & 74.55 & 78.26 & 81.52 & 79.34 & 85.04 & 86.30 & 75.78 & 84.23 & 82.73 \\
Mistral-embed & 84.73 & 77.09 & 83.48 & 73.48 & 80.94 & 82.13 & 79.08 & 79.38 & 80.81 & 72.85 & 78.74 \\
Ada-002 & 84.44 & 79.50 & 82.14 & 72.75 & 81.29 & 82.30 & 80.76 & 80.08 & 80.10 & 73.01 & 79.23 \\ \bottomrule
\end{tabular}
\caption{Retrieval performance (in recall@1 \%) of retrieval approaches in different subjects and question types.}
\label{tab:retrieval_result}
\end{table*}
\vspace{0.2cm}
\begin{table*}[h!]
\centering
\scriptsize
\begin{tabular}{lrrrrrcccccc}
\toprule
\multicolumn{1}{c}{\multirow{2}{*}{\textbf{Large Language Models}}} & \multicolumn{5}{c}{\textbf{Subjects}} & \multicolumn{5}{c}{\textbf{Question Types}} & \multirow{2}{*}{\textbf{Average}} \\ \cmidrule(lr){2-11}
\multicolumn{1}{c}{} & \multicolumn{1}{c}{\textit{chem.}} & \multicolumn{1}{c}{\textit{bio.}} & \multicolumn{1}{c}{\textit{phys.}} & \multicolumn{1}{c}{\textit{geo.}} & \multicolumn{1}{c}{\textit{hist.}} & \multicolumn{1}{l}{\textit{Sim. Dir.}} & \multicolumn{1}{l}{\textit{Mul. Dir.}} & \multicolumn{1}{l}{\textit{Mul. Dis.}} & \multicolumn{1}{l}{\textit{Mul. Imp.}} & \multicolumn{1}{l}{\textit{Dis. Imp.}}  &  \\ \midrule
Llama3-8b & 77.81 & 75.60 & 75.45 & 80.14 & 78.65 & 90.33 & 88.85 & 89.69 & 63.55 & 49.43 & 77.77 \\
GPT-3.5-turbo & 78.93 & 80.53 & 82.59 & 83.04 & 82.43 & 92.62 & 90.53 & 91.24 & 71.22 & 57.72 & 81.73 \\
GPT-4-turbo & 85.30 & 89.23 & 86.16 & 90.00 & 90.13 & 95.74 & 96.18 & 96.19 & 81.06 & 71.71 & 88.99 \\ \bottomrule
\end{tabular}
\caption{Performance (in accuracy \%) of LLMs in question answering in different subjects and question types.}
\label{tab:llm_full_result}
\end{table*}

\newpage
\section{Prompting Approach}
\label{app:prompting_approach}
In our experiments, we adopt the prompting approach of \textit{Locate-and-Answer}, to facilitate active acquisition of information in the context when performing question answering. We first request LLMs to identify and locate the corresponding sentence that include the knowledge for the question from the provided document, and then reason to provide its answer. According to the experimental result in Table \ref{tab:llm_prompting}, this prompting approach can effectively improve the QA performance of LLMs compared to direct answering.
\begin{table*}[h]
\centering
\scriptsize
\begin{tabular}{lcc}
\toprule
\multicolumn{1}{c}{\multirow{2}{*}{\textbf{Model}}} & \multicolumn{2}{c}{\textbf{Prompting Method}} \\\cmidrule{2-3} 
\multicolumn{1}{c}{} & \multicolumn{1}{l}{Direct Answer} & Locate-and-Answer \\ \midrule
Gemini-1.5-flash & 75.21 & 78.54 \\
GPT-3.5-turbo  & 73.91 & 81.73 \\
GPT-4o &  80.43 & 87.09 \\ \bottomrule
\end{tabular}
\caption{Performance (in accuracy \%) of LLMs in question answering with different prompting methods.}
\label{tab:llm_prompting}
\end{table*}

\section{Calibration, Parametric Knowledge, and Instruction Following in RAG Systems}
\label{app:calibration-induced_performance_discrepancies}

In our study, retrieved documents serve as the unequivocal reference, mirroring the pedagogical approach in K-12 education where textbooks are authoritative.
However, our evaluation results reveal counterintuitive performance patterns in state-of-the-art LLMs, particularly concerning simple direct questions.
Stronger models, such as GPT-4 and GPT-4o, occasionally underperform compared to weaker ones on these questions, which primarily assess direct factual recall from the provided (and hypothetically updated) context.
This suggests that the calibration of these advanced LLMs might lead them to exhibit excessive confidence in their internal, parametric knowledge, thereby inhibiting their consistent reliance on the external documents explicitly provided as the ground truth for the task.

This observation highlights a critical tension inherent in RAG systems: the trade-off between an LLM's adherence to its ingrained parametric knowledge (what might be termed its internal "factuality calibration") and its capacity for robust instruction following.
While an LLM prioritizing its internal knowledge, especially if it aligns with generally accepted facts, could be interpreted as a form of "honesty," this perspective shifts within the specific operational context of RAG systems like those evaluated with \textsc{KnowShiftQA}.
In such systems, LLMs are explicitly instructed to base their responses on the provided contextual documents.
Therefore, when a model disregards this instruction and favors its parametric knowledge---particularly when the context contains deliberate, hypothetical updates---it signifies a limitation in its instruction-following capability rather than a commendable adherence to broader factual accuracy.

The underperformance on simple direct questions can thus be partly attributed to this conflict: the models may struggle to override a strong internal "prior" when explicitly instructed to use new, conflicting information from the context.
This distinction is paramount for developing robust educational QA systems.
Such systems must reliably reflect the content of designated authoritative sources (e.g., textbooks), even if these sources present information that has been updated or differs from an LLM's training data.
Consequently, the ability to flexibly integrate and prioritize instructed contextual information is a fundamental requirement for their effectiveness in dynamic knowledge environments.
While in real-world scenarios with less clearly defined ground truths or more ambiguous knowledge conflicts, the sophisticated reasoning of stronger LLMs might offer advantages in navigating discrepancies, the specific demands of an educational RAG system underscore the critical importance of faithful instruction adherence.
\end{document}